\documentclass[letterpaper]{article} 
\usepackage{aaai2026}  
\usepackage{times}  
\usepackage{helvet}  
\usepackage{courier}  
\usepackage[hyphens]{url}  
\usepackage{graphicx} 
\usepackage{natbib}  
\usepackage{caption} 
\usepackage{algorithm}
\usepackage{algorithmic}

\usepackage{comment}
\usepackage{pdfpages}
\usepackage{xcolor}
\usepackage[framemethod=tikz]{mdframed}
\usetikzlibrary{shadows}
\newmdenv[shadow=true,shadowcolor=black,rightmargin=8pt]{shadedbox}

\newboolean{draft_comments}
\setboolean{draft_comments}{false}
\ifthenelse{\boolean{draft_comments}}
{
  \newcommand{\Alan}[1]{\textcolor{purple}{[ALAN: #1]}}
  \newcommand{\MichaelD}[1]{\textcolor{blue}{[MICHAELD: #1]}}
  \newcommand{\Brian}[1]{\textcolor{cyan}{[BRIAN: #1]}}
  \newcommand{\MichaelM}[1]{\textcolor{teal}{[MICHAELM: #1]}}
  \newcommand{\Reed}[1]{\textcolor{brown}{[REED: #1]}}
  \newcommand{\reminder}[1]{\textcolor{green}{\textit{\textsc{#1}}}}
}
{
  \newcommand{\Alan}[1]{}
  \newcommand{\MichaelD}[1]{}
  \newcommand{\Brian}[1]{}
  \newcommand{\MichaelM}[1]{}
  \newcommand{\Reed}[1]{}
  \newcommand{\reminder}[1]{}
}
\usepackage{fancyhdr}
\fancypagestyle{firstpage}{
  \fancyhf{}
  \fancyhead[R]{SAND2025-15437A}
  
}
\usepackage{newfloat}
\usepackage{listings}
\DeclareCaptionStyle{ruled}{labelfont=normalfont,labelsep=colon,strut=off} 
\floatstyle{ruled}
\newfloat{listing}{tb}{lst}{}
\floatname{listing}{Listing}
\title{Toward Maturity-Based Certification of Embodied AI: \\Quantifying Trustworthiness Through Measurement Mechanisms}
\author{Michael C. Darling, Alan H. Hesu, Michael A. Mardikes, Brian C. McGuigan, Reed M. Milewicz}

\affiliations{
    Sandia National Laboratories, Livermore CA \& Albuquerque NM, USA\\
    \texttt{\{mcdarli, ahhesu, mamardi, bcmcgui, rmilewi\}@sandia.gov}
}
    
\date{\today}

\begin{document}

\maketitle

\begin{abstract}
We propose a maturity-based framework for certifying embodied AI systems through explicit measurement mechanisms. We argue that certifiable embodied AI requires structured assessment frameworks, quantitative scoring mechanisms, and methods for navigating multi-objective trade-offs inherent in trustworthiness evaluation. We demonstrate this approach using uncertainty quantification as an exemplar measurement mechanism and illustrate feasibility through an Uncrewed Aircraft System (UAS) detection case study.
\end{abstract}

\section{The Certification Challenge for Embodied AI}
\thispagestyle{firstpage}
Embodied AI (E-AI) refers to AI systems with a physical presence, such as autonomous vehicles, drones, or healthcare devices, which can perceive and act in the physical world. Certification of trustworthiness must be required to deploy E-AI systems in safety-critical contexts.

E-AI systems must be able to respond to unpredictably changing physical environments, be reliable and robust both in terms of hardware and software (such as sensor failures and misinterpretations), be able to respond to dynamic situations while still being sufficiently predictable and transparent to human actors. They must also be resilient to cyber-kinetic insults that go beyond the typical security threats to AI systems. While verification techniques and trustworthiness frameworks are advancing, the state of the art is not keeping pace with the emerging challenges of E-AI systems.

Current verification approaches in the AI community often focus on establishing specific, measurable properties in isolation: improving accuracy on held-out test sets, demonstrating robustness to adversarial perturbations, or ensuring fairness across demographic groups. However, certification requires holistic assessment of trustworthiness throughout the development lifecycle from requirements specification through deployment and runtime monitoring. This is especially true for E-AI systems, which require a continuous renewal of trust as they act and respond to real-world conditions over extended periods of operation; this necessitates a whole-of-system approach to addressing trustworthiness.

The NIST AI Risk Management Framework defines characteristics of trustworthy AI: valid and reliable, safe, secure and resilient, accountable and transparent, explainable and interpretable, privacy-enhanced, and fair with harmful bias managed~\cite{nist_ai_rmf}. While these characteristics provide goal posts, operationalizing them into measurable, auditable criteria that can support certification decisions remains an open challenge. For E-AI systems, there is also an added layer of translation and friction for developers and users, as concepts and tactics which work for trustworthiness in more conventional AI contexts do not necessarily map onto E-AI systems; interpretability, for instance, takes on new dimensions in human-autonomy teaming as human and AI actors must interpret each other's movements in real time.

We argue that certifiable E-AI requires maturity models that provide structured assessment frameworks, quantitative scoring mechanisms, and explicit methods for navigating multi-objective trade-offs inherent in trustworthiness evaluation. Furthermore, we propose that explicit, quantifiable measurement mechanisms can operationalize abstract trustworthiness characteristics into concrete evidence for verification. We argue that this would bring much needed clarity to the design and use of these complex systems. We illustrate this approach through the lens of uncertainty quantification (UQ) as an exemplar measurement mechanism and explore feasibility with an Uncrewed Aircraft System (UAS) detection system example~\cite{wang2021uas_variability}.

\section{Proposed Framework: Maturity-Based Trustworthiness Assessment}

Maturity models, such as the Capability Maturity Model Integration (CMMI), have proven effective for assessing and improving software development processes~\cite{cmmi_dev}. We envision a maturity model approach tailored to the unique challenges of E-AI systems. The framework we propose would comprise three interconnected components:

\textbf{Dimensional Assessment Structure:} Map NIST trustworthiness characteristics to stages of the ML development lifecycle (requirements, data collection/curation, model training, validation/testing, deployment, monitoring). Each intersection represents an assessable element. For example, ``robustness at the testing stage'' or ``privacy at the deployment stage.'' This creates a structured matrix for comprehensive trustworthiness evaluation.

\textbf{Maturity Scoring Methodology:} Define maturity levels for each trustworthiness characteristic, with level-specific criteria and required evidence. Many maturity models use five levels ranging from Initial to Optimizing. For AI trustworthiness, we envision a similar structure. A notional example for robustness:

\begin{itemize}
\item \textbf{Robustness Level 1 (Ad-hoc Testing):} Limited scenario testing with informal robustness claims. 

\item \textbf{Robustness Level 2 (Structured Testing):} Documented test scenarios covering identified operational conditions. 

\item \textbf{Robustness Level 3 (Measurement-Driven):} Systematic testing with measurement-guided scenario generation and quantified performance bounds. 

\item \textbf{Robustness Level 4 (Statistical Guarantees):} Formal statistical guarantees (such as conformal prediction~\cite{shafer2008tutorial} with specified coverage) validated across operational domain. Runtime monitoring with measurement-based triggers.

\item \textbf{Robustness Level 5 (Formal Verification):} Mathematical proofs about system components that combine to guarantee whole-system properties. Runtime monitoring with formally verified safety mechanisms ensuring guaranteed responses to violations.
\end{itemize}

Critically, each maturity level must specify what evidence constitutes achievement.

\textbf{Multi-Objective Optimization:} Some trustworthiness characteristics inherently trade off against each other and against performance objectives. Transparency mechanisms may reduce accuracy; privacy preservation may limit explainability. Rather than treating these as ad-hoc engineering compromises, we propose using multi-objective optimization to make trade-offs explicit, quantifiable, and defensible in certification contexts~\cite{marler2004survey}.

\subsection{Measurement Mechanisms with UQ Exemplar}

Abstract trustworthiness principles must be operationalized through explicit, quantifiable measurement mechanisms that can produce verifiable evidence throughout the certification process. Existing maturity models for AI trustworthiness, such as MM4XAI-AE for explainability, rely on binary indicators assessed through documentation review: an approach well-suited for retrospective audits but insufficient for safety-critical embodied systems requiring continuous, runtime-integrated assessment~\cite{munoz2025maturity}.  We propose that effective measurement mechanisms share four critical properties:

\begin{enumerate}
\item \textbf{Quantifiable Metrics:} The mechanism must produce numerical measurements with clear thresholds that can define maturity level boundaries.

\item \textbf{Actionable Outputs:} The mechanism must produce outputs that directly connect to concrete system decisions and safety mechanisms. Measurements should trigger specific actions such as alerting operators or invoking fallback procedures. This transforms measurements from diagnostic information into active components of trustworthy system behavior.

\item \textbf{Formal Properties:} Where possible, the mechanism should provide mathematical guarantees that support verification. 

\end{enumerate}

Comprehensive certification requires that measurement mechanisms collectively provide coverage across the development and operational lifecycle, from requirements specification through runtime monitoring. Individual mechanisms may be most applicable at specific stages; the certification process must integrate multiple mechanisms to achieve full lifecycle coverage.

\subsubsection{UQ Demonstration of the Four Properties}

UQ provides an example of how measurement mechanisms satisfy the required properties:

\textbf{Quantifiable metrics:} UQ techniques provide numerical outputs including calibration error~\cite{guo2017calibration}, entropy~\cite{kendall2017uncertainties}, ensemble variance~\cite{lakshminarayanan2017simple}, out-of-distirbution detection scores~\cite{hendrycks2016baseline}, and conformal prediction set sizes~\cite{shafer2008tutorial}. Thresholds of these metrics could directly map to maturity levels:

\textbf{Actionable outputs:} UQ measurements directly drive system decisions and safety mechanisms. When uncertainty exceeds thresholds, the system can, for examples, request human review or switch to conservative fallback behaviors.

\textbf{Formal properties:} UQ encompasses methods with varying degrees of mathematical rigor. While softmax confidence scores provide only heuristic uncertainty estimates,  more sophisticated methods, such as conformal prediction, can provide guarantees on prediction set coverage. Notably, UQ quality can differ amongst methods highlighting the need for principled  assessment~\cite{adams2023improving}.

\textbf{Lifecycle integration:} UQ demonstrates applicability across the full embodied AI lifecycle:

\begin{itemize}
\item \textit{Requirements phase:} UQ informs operational domain specifications (``system must maintain uncertainty below 0.3 in specified weather conditions'') and sensor selection criteria

\item \textit{Data collection phase:} Uncertainty measurements identify data gaps, enabling active learning and targeted data acquisition. 

\item \textit{Training phase:} UQ considerations influence architecture choices (ensembles vs. single models), loss function design (incorporating calibration objectives), and regularization strategies.

\item \textit{Validation/testing phase:} Calibration metrics, OOD detection performance, and coverage validation provide quantifiable criteria that can define pass/fail thresholds for maturity assessment.

\item \textit{Integration phase:}  Uncertainty propagation through system components reveals how model uncertainty affects end-to-end system behavior. For systems integrating multiple sensors, data-driven UQ methods can quantify how uncertainties from individual sources combine in downstream analytics~\cite{stracuzzi2018data}.

\item \textit{Deployment phase:} Real-time uncertainty estimates enable runtime monitoring and threshold-based guardrails.

\item \textit{Operations/maintenance phase:} Longitudinal uncertainty tracking detects performance degradation, distribution shift, and anomalies that may indicate sensor degradation or hardware issues.
\end{itemize}

This integration across both ML and physical system lifecycles is particularly critical for embodied AI, where sensor degradation can cause distribution shift.

\subsubsection{Connecting UQ to NIST Characteristics}

UQ principles extend across multiple NIST trustworthiness characteristics.

\textbf{Reliability/Validity:} Probability calibration measures whether a model's predicted confidences match its actual frequency of being correct.

\textbf{Robustness:} Out-of-distribution (OOD) detection identifies when a model encounters examples beyond its training distribution.

\textbf{Transparency/Explainability:} Transparency/Explainability: Uncertainty estimates can identify regions of the input space where predictions are credible versus regions requiring further analysis~\cite{darling2019using}.  Uncertainty decomposition into epistemic (model ignorance, reducible with more data) versus aleatoric (inherent data noise, irreducible) components provides interpretable confidence explanations. 

\textbf{Safety:} Uncertainty thresholds could trigger runtime guardrails, preventing unsafe actions. 

While UQ addresses several trustworthiness characteristics, comprehensive assessment requires complementary measurement mechanisms.

\subsubsection{Open Research Questions}

The UQ exemplar raises questions that generalize across measurement mechanisms:

\textbf{Mechanism development:} What measurement mechanisms are appropriate for each NIST characteristic, and which existing techniques from ML research, formal methods, or software engineering can be adapted? 

\textbf{Maturity mapping:} How do we ensure maturity levels are comparable across different trustworthiness characteristics?

\textbf{Evidence sufficiency:} What combinations of mechanisms provide sufficient evidence for certification decisions? 

\textbf{Lifecycle tooling:} How do we integrate multiple measurement mechanisms into existing development workflows without overwhelming developers?

\textbf{Physical-software integration:} For embodied AI specifically, how do measurement mechanisms account for hardware-software coupling?

\section{UAS Detection: A Motivating Case Study}
\label{sec:uas}

Our ongoing work in UAS detection exemplifies both the necessity and feasibility of this approach, using uncertainty quantification as the exemplar measurement mechanism. UAS detection systems represent safety-critical embodied AI where failure modes have significant consequences.

UAS detection represents embodied AI since these systems integrate physical sensors (radar, RF receivers, cameras, acoustic arrays) mounted on physical platforms (fixed installations, mobile vehicles, or counter-UAS drones) that must perceive and respond to physical threats (incoming drones) in real-world environments. The AI component processes sensor data to detect, classify, and track physical objects, and its outputs drive physical responses: alerting human operators, triggering tracking systems, or activating countermeasures. 

The trustworthiness challenges are inherently embodied: sensor degradation affects ML performance, environmental conditions (weather, terrain, electromagnetic interference) impact both sensing and inference, and the consequences of decisions manifest physically (such as allowing a drone to penetrate restricted airspace). Beyond the uncertainties inherent to all AI systems, UAS detection must maintain trustworthiness under uncertainties stemming from hardware.
\Alan{
notes
- disagreement between multiple sensors, eg. two cameras, or camera + radar. This gets into some of the downstream impacts, like doing data fusion of multiple sensor streams. Maybe some crossover with ensembling?
- coupled v&v challenges - UQ and other metrics depend on how error is propogated from things like sensor uncertainty, so how do we quantify that stuff too? On the other hand, doing this sort of system characterization and modeling may also enable the use of UQ to sense when hardware or other environmental conditions have degraded
- system adaptability/reconfigurability - in operation, we may want to reconfigure or redeploy - for example, move where cameras are pointed, add new sensors, deploy in a new location, etc. Understanding the system well enables making these decisions and understanding their ramifications, eg. how additional sensors affect uncertainty
}

The safety-criticality stems from asymmetric failure costs. False negatives (missed detections) enable security threats. Conversely, false positives create multiple problems. In physical security contexts, high false alarm rates (FAR) or nuisance alarm rates (NAR) degrade human operator vigilance and trust in the system.  Operators become desensitized to alerts and may ignore genuine threats \cite{cvach2012alarm}. The multi-objective challenge of balancing security (minimize false negatives), operator trust (minimize false alarms) exemplifies why measurement mechanisms and maturity-based frameworks are essential for navigating complex trustworthiness trade-offs.

\subsection{The Verification Challenge}

UAS detection must demonstrate robustness across enormous variability \cite{wang2021uas_variability, wilson2020uas_rand}: different aircraft types and sizes, varied geographic terrains (urban, forested, maritime, desert), lighting conditions (dawn, dusk, direct sunlight, overcast), weather conditions, viewing angles, and crucially, adversarial modifications to UAS appearance. As UAS usage is expected to increase in the private sector, detection is increasingly relevant to many civilian contexts; This includes preventing errant UASs from unwittingly entering restricted spaces such as near airports as well as intercepting unauthorized UASs being used to harass or disrupt operations. The same concerns exist in military operations as well (such as battlefields in which both sides have their own deployed drone fleets flying in every direction). In all these cases, real-world data collection across this scenario space is expensive and time-consuming \cite{brewczynski2024methods}.

This verification challenge illustrates why measurement mechanisms are essential: we need quantifiable ways to assess whether testing coverage is adequate, whether the system knows when it's uncertain, and whether robustness claims are justified.

\subsection{Closed-Loop Synthetic Data Generation}
\Alan{
more details on our approach (see paragraph below)
- 3D rendering of assets and environments for synthetic data generation using Unreal Engine
- image-based deep learning models for UAS detection (\cite{sahay_uncertainty_2022})
- ensemble-based UQ metrics (cite stuff here?)
- generating synthetic data:
 - measure uncertainty of samples in validation dataset and generate additional synthetic data that is similar (w.r.t. sample generation parameters) to the most uncertain samples
 - measure similarity in latent space using UMAP (\cite{mcinnes_umap_2020}) in feature space using sample generation parameters

}

We are developing a synthetic data pipeline that enables systematic data generation guided by uncertainty analyses. The pipeline not only addresses the real-world data collection burden but also provides external control over critical parameters including UAS characteristics (type, size, pose, appearance, adversarial modifications), environmental factors (geographic location, terrain type, time of day, weather), confounding factors (birds and clutter objects)

We leverage this pipeline in a closed loop fashion to characterize and improve image-based deep learning models for UAS detection (\cite{sahay_uncertainty_2022}). By using primarily ensemble-based methods for measuring uncertainty, we discover potential robustness gaps. We then address these gaps by generating additional synthetic data that is similar to prior samples with high uncertainty, retrain the models, and reassess uncertainty. Similarity is measured in the latent space using uniform manifold approximation and projection (UMAP)~\cite{mcinnes_umap_2020} and in the feature space via the UAS characteristics, environmental factors, and other synthetic sample generation parameters.

Synthetic data generation has become increasingly sophisticated, with methods ranging from generative adversarial networks to physics-based simulation \cite{de2022next, paulin2023review}. These techniques enable creation of diverse, realistic training and testing scenarios while maintaining precise control over parameters for systematic robustness assessment.

The closed-loop approach demonstrates how measurement mechanisms can actively guide system improvement, not just passively assess it. 

\subsection{Preliminary Findings and Open Questions}

Our preliminary results demonstrate a correlation between prediction uncertainty and classification error: the model is less likely to be correct when uncertainty is high. This indicates that measurement mechanisms like UQ can serve trustworthiness assessment.

However, this finding immediately raises critical questions for maturity-based certification:

\textbf{Threshold determination:} At what uncertainty level should the system trigger alerts, refuse to decide, or invoke fallback mechanisms? How do we set these thresholds for different deployment contexts (military vs. civilian airspace)? Runtime assurance frameworks have explored similar questions for safety-critical control systems, but extending these concepts to ML-based perception systems and mapping them to maturity levels remains unexplored. 

\textbf{Feature attribution:} We observe uncertainty patterns but have not yet identified which specific features (such as lighting, aircraft size, terrain complexity) drive uncertainty. Understanding these relationships is essential for requirements specification and test coverage assessment. 

\textbf{Maturity scoring:} How do we translate ``model shows high uncertainty in forested terrain at dusk'' into a quantitative robustness maturity score? 

\textbf{Multi-objective trade-offs:} Detection sensitivity vs. false alarm rate illustrates a classic trade-off with trustworthiness implications. False negatives threaten security (safety/reliability concern) while false positives threaten operator trust (human factors concern). How do we formalize this trade-off for certification decisions? 

\textbf{Test adequacy:} How much synthetic data generation and testing is ``enough'' to claim adequate scenario coverage? Can we develop formal coverage metrics analogous to code coverage in software testing?

\section{Research Agenda}

We identify a research direction and  critical open problems:

\textbf{Maturity Model Design:}

\begin{itemize}
\item What maturity level structure makes sense for embodied AI systems? 
\item How do we design maturity criteria that drive meaningful improvement rather than ``checkbox compliance''?

\end{itemize}

\textbf{Measurement Methodology:}

\begin{itemize}
\item Which measurement mechanisms are most mature and suitable for each NIST characteristic?
\item What combination of testing, formal verification, and runtime monitoring constitutes sufficient evidence for each maturity level?

\end{itemize}

\textbf{Multi-Objective Optimization Formalization:}

\begin{itemize}
\item How do we mathematically represent trade-offs between trustworthiness characteristics measured by different mechanisms?
\item What decision-theoretic frameworks can support stakeholders in navigating design trade-offs and certification authorities in setting appropriate standards? Preliminary work has explored formal methods for linking ML outputs to optimal decisions under uncertainty~\cite{field2022decision}, but extension to multi-objective trustworthiness trade-offs remains open.

\end{itemize}

\textbf{Integration with Formal Methods:}

\begin{itemize}
\item How do measurement-based maturity assessments connect to formal verification techniques?
\item Which measurement mechanisms can provide formal guarantees and how do we prioritize these for high-maturity certification?
\end{itemize}

\textbf{Runtime Assurance:}

\begin{itemize}
\item How do maturity assessments translate into runtime monitoring requirements?
\item What guardrails should measurement mechanism outputs trigger (alerts, fallbacks, conservative actions)?
\item How do we validate that runtime monitors themselves are trustworthy?
\end{itemize}

\section{Conclusion}

The path to certifiably trustworthy embodied AI requires structured frameworks that connect abstract trustworthiness principles to concrete, measurable evidence throughout the development lifecycle, not just verification techniques in isolation.  We argue that maturity models, operationalized through explicit measurement mechanisms, offer a promising direction. Uncertainty quantification demonstrates the feasibility of this approach, and we invite the community to help extend it across all trustworthiness characteristics.

\section{Acknowledgements}
Sandia National Laboratories is a multi-mission laboratory managed and operated by National Technology \& Engineering Solutions of Sandia, LLC (NTESS), a wholly owned subsidiary of Honeywell International Inc., for the U.S. Department of Energy’s National Nuclear Security Administration (DOE/NNSA) under contract DE-NA0003525. This written work is authored by an employee of NTESS. The employee, not NTESS, owns the right, title and interest in and to the written work and is responsible for its contents. Any subjective views or opinions that might be expressed in the written work do not necessarily represent the views of the U.S. Government. The publisher acknowledges that the U.S. Government retains a non-exclusive, paid-up, irrevocable, world-wide license to publish or reproduce the published form of this written work or allow others to do so, for U.S. Government purposes. The DOE will provide public access to results of federally sponsored research in accordance with the DOE Public Access Plan.

\bibliography{references}

\end{document}